\pdfoutput=1
\documentclass[11pt]{article}

\newcommand*{\fixmeON}{} % Comment out this line to turn FIXMEs off
\newcommand{\covid}[1]{COVID-19}

\newcommand{\fbert}[1]{$F_{BERT}$ score}
\newcommand{\LSTM}[1]{{%
\texttt{LSTM}}}

\newcommand{\CV}[1]{{%
\texttt{CV}}}

\newcommand{\CVlong}[1]{{%
{CountVectorizer}}}

\newcommand{\WE}[1]{{%
\texttt{WE}}}

\newcommand{\WElong}[1]{{%
{Word Embeddings}}}

\newcommand{\TFIDF}[1]{{%
\texttt{TF-IDF}}}

\newcommand{\roberta}[1]{\texttt{RoBERTa}}
\newcommand{\privbert}[1]{\texttt{PrivBERT}}
\newcommand{\svm}[1]{\texttt{SVM}}
\newcommand{\rf}[1]{\texttt{RF}}

\newcommand{\REDACT}[1]{$\Box REDACTED \Box$} 

\newcommand{\redactCollege}[1]{[a U.S. University]}  

\newcounter{boldifyCounter}
\newcounter{fixmeSectionCounter}
\newcounter{fixmeTotalCounter}
\makeatletter
\@addtoreset{fixmeSectionCounter}{section}
\@addtoreset{fixmeSectionCounter}{subsection}
\@addtoreset{boldifyCounter}{section}
\@addtoreset{boldifyCounter}{subsection}
\makeatother

\newcommand{\boldify}[1]{}% boldify takes 1 argument and ignores it when boldifies are OFF
\ifdefined\boldifyON
	\renewcommand{\boldify}[1]{% ... and it does more when they are on
        \par\noindent%  suppress the indent for this new paragraph
		\stepcounter{boldifyCounter}% we are handling a boldify now, so we need to increment
		\textbf{{\color{green}**}% get the nice green stars we love
		~\arabic{section}.\arabic{subsection}.\arabic{boldifyCounter}% grab the current section numbers and format them as arabic
		: #1} % and finally the boldify itself
	}
\fi % end the boldify if. Next line tells the word counter to IGNORE all text in this command (it is hidden!)
\newcommand{\reportOnFIXME}{%
    \newcount\iterCounter
    \iterCounter=1
    \newcount\endCounter
    \endCounter=\totvalue{fixmeTotalCounter}
    \advance \endCounter +1
    There are 
    {\color{red}\total{fixmeTotalCounter}} 
    FIX\_ME\\
    links:
    \loop
        \hyperlink{fixTag\the\iterCounter}{\#\the\iterCounter}
        \advance \iterCounter +1
    \ifnum \iterCounter < \endCounter
    \repeat
}

\newcommand{\FIXME}[1]{} % FIXME takes 1 argument and ignores it when FIXME is OFF
\ifdefined\fixmeON
	\renewcommand{\FIXME}[1]{\par\noindent% ... and it does more when they are on
		\stepcounter{fixmeSectionCounter}\stepcounter{fixmeTotalCounter}% we are handling a FIXME, need to increment
		{\color{red}\fbox{\color{black}% get that nice red framebox that we love, and switch the text back to black
			\parbox{.965\linewidth}{% put the text inside an invisible box of this size
				\textbf{\hypertarget{fixTag\thefixmeTotalCounter}{FIXME}	\arabic{section}.\arabic{subsection}.% get the section numbers and format as arabic
        		\arabic{fixmeSectionCounter} (\color{red}% switch to red to show the total counter
        		\#\arabic{fixmeTotalCounter}):} #1}}% and then finally the FIXME text itself
        }
	}
\fi % end the FIXME if. Next line tells the word counter to IGNORE all text in this command (it is hidden!)
\newcommand{\FIXED}[1]{}
\ifdefined\fixmeON
	\renewcommand{\FIXED}[1]{\par\noindent%
		{\color{black}\fbox{\color{black}%
			\parbox{.99\columnwidth}{%
				\color{blue}#1}}%
        }
	}
\fi % end the FIXED if. Next line tells the word counter to IGNORE all text in this command (it is hidden!)
\newcommand{\draftStatus}[1]{}% draftStatus takes 2 argument and ignores it when draft statuses are OFF
\ifdefined\draftStatusON
	\renewcommand{\draftStatus}[1]{% ... and it does more when they are on
        \hfill **#1
	}
\fi % end the draft status if. Next line tells the word counter to IGNORE all text in this command (it is hidden!)
\usepackage{totcount}\regtotcounter{fixmeTotalCounter}
\usepackage{datetime2} % for versioning info
\usepackage{enumitem}
\usepackage{url}
\usepackage{hyperref}
\usepackage{twocolceurws}

\usepackage{times}
\usepackage{latexsym}

\usepackage[T1]{fontenc}
\usepackage[utf8]{inputenc}

\usepackage{microtype}

\usepackage{inconsolata}

\usepackage{graphicx}

\title{Fake News Detection After LLM Laundering: Measurement and Explanation}

\author{
Rupak Kumar Das \\ College of IST\\
Pennsylvania State University\\
                PA, 16801, USA \\ rjd6099@psu.edu
\and
Dr. Jonathan Dodge \\ College of IST\\
Pennsylvania State University\\
                PA, 16801, USA \\ jxd6067@psu.edu}

\newcommand{\theAbstract}[1]{%
With their advanced capabilities, Large Language Models (LLMs) can generate highly convincing and contextually relevant fake news, which can contribute to disseminating misinformation.
Though there is much research on fake news detection for human-written text, the field of detecting LLM-generated fake news is still under-explored. 
This research measures the efficacy of detectors in identifying LLM-paraphrased fake news, in particular, determining whether adding a paraphrase step in the detection pipeline \textit{helps} or \textit{impedes} detection.
This study contributes:
(1)~Detectors struggle to detect LLM-paraphrased fake news more than human-written text,
(2)~We find which models excel at which tasks (evading detection, paraphrasing to evade detection, and paraphrasing for semantic similarity).
(3)~Via LIME explanations, we discovered a possible reason for detection failures: sentiment shift.
(4)~We discover a worrisome trend for paraphrase quality measurement: samples that exhibit sentiment shift \textit{despite} a high BERTSCORE.
(5)~We provide a pair of datasets augmenting existing datasets with paraphrase outputs and scores. The dataset is available on GitHub\footnote{\url{https://github.com/rupakdas18/Fake-News-Detection-After-LLM-Laundering}}.

}

\institution{}

\begin{document}
%%%%%%%%%%%%%%%%%%%%%%%%%%
%\input{0-TODO_LIST}
%%%%%%%%%%%%%%%%%%%%%%%%

\maketitle

\begin{abstract}
\theAbstract{}
\end{abstract}

%\clearpage
\section{Introduction
\draftStatus{APPROVED}}

\boldify{What is paraphrasing and its importance}

Paraphrasing is the process of generating text from a reference text with syntactic and lexical diversity while maintaining semantic similarity. 
Paraphrasing is important for different downstream NLP tasks, such as text summarization~\cite{cao2017joint}, semantic parsing~\cite{berant2014semantic}, question answering~\cite{fader2014open, yin2015answering}, data augmentation~\cite{yu2018qanet}, adversarial example generation~\cite{iyyer2018adversarial}, and checking the robustness of a model~\cite{iyyer2018adversarial}.
However, effective paraphrasing is challenging because it involves syntactically rephrasing text, but preserving meaning~\cite{khurana2023natural}. 

\boldify{Paraphrasing in fake news detection:
This is an under-explored aspect of a VERY important topic.}

The impact of paraphrasing on fake news detection is still under-explored, and the advancement of large language models only increases the importance of this field.
OpenAI reported ongoing attempts to misuse AI for political misinformation.
Still, the most widespread incident was a hoax falsely claiming to involve its models, with the overall impact on the 2024 election appearing modest~\cite {openAIreport}.

\boldify{Paraphrasing in fake news detection: What do SOTA detectors do? and how does LLM paraphrasing factor into the picture?}

State-of-the-art fake news detectors mainly distinguish real from fake based on human knowledge (expert- or crowdsourcing-oriented), content features (linguistic, syntactic, and sentiment), and network features.
Those features may make it easier for state-of-the-art detectors to detect LLM-generated/synthesized fake news because of their patterns of generating fake news.

\boldify{What LLMs can do to impact HUMANS through generating or paraphrasing fake news can impact us? Why they are so dangerous.}

The impact of the generated text by different LLMs (e.g., GPT~\cite{brown2020language}, BERT~\cite{devlin2018bert}, T5~\cite{raffel2020exploring}, LLaMA~\cite{touvron2023llama}) on fake news detection systems has recently attracted attention.
LLMs are reasonably good at generating and synthesizing fake news.
LLMs like GPT2 can synthesize and spread misinformation by pre-training it to a large-scale news corpus~\cite{zellers2019defending}.
Further, LLM-generated fake news is more controllable because the generation process conditions on knowledge elements (entities, relations, and events) taken from the original news article~\cite{huang2022faking}.
Language models can generate paraphrased text by using `deceptive style'~\cite {chen2023can} to generate fake news, making it difficult for state-of-the-art detectors to detect. 
A few recent studies explored how fake news detectors react to LLM-generated text.
\cite{su2023fake} demonstrated that fake news detectors frequently mistakenly authenticate human-written fake news, but are more likely to identify LLM-generated content as fake news because of the implicit linguistic patterns of LLM outputs.
Similarly, LLM transformers from the BERT family are more successful in classifying GPT-generated articles than non-GPT-generated fake news~\cite{stewart2023efficacy}.
However, \cite{chen2023can} find that compared to human-written misinformation with the same semantics, LLM-generated misinformation may be more difficult for humans and detectors to identify because of LLMs' deceptive styles.
Moreover, LLM-based fake news detectors also struggle with self-generated fake news~\cite{jiang2023disinformation}.

\boldify{why we need to evaluate the paraphrased text}

A good paraphraser conveys the same semantics and keeps the sentence grammatically correct.
Evaluating the paraphrased text is crucial because any shift in semantics can create problems in detecting fake news. 
Previously, researchers used BLEU~\cite{papineni2002bleu} and ROUGE~\cite{lin2004rouge} to compare a pair of sentences. 
Those methods use the n-gram technique to determine the similarity between a reference and a candidate sentence.
However, this type of n-gram technique fails to match paraphrased texts correctly~\cite{zhang2019bertscore}.
Researchers use another technique called TER~\cite{snover2006study}, which finds the minimum number of actions required to get the reference sentence from the candidate sentence.
However, TER focuses more on word matching than semantic similarity~\cite{lee2023survey}. 
Word embeddings-based techniques such as MEANT~\cite{lo2017meant} methods are also prevalent, but word embedding does not consider the surrounding words while representing the words as word vectors. 
Instead, contextual embedding-based techniques such as BERTSCORE~\cite{zhang2019bertscore} better measure semantic similarity.

\boldify{Why this research?}

Measuring the efficacy of detectors in distinguishing fake news produced by LLMs contributes to the welfare of society in the following ways.
First, the investigation clarifies whether adding a paraphrase step to the detection pipeline helps or impedes the process.
Answering this question allows us to improve techniques to combat disinformation by identifying an attack or a defense.
Second, assuming that paraphrasing \textit{helps} detection (defense mode), our research clarifies the best paraphrasing method to incorporate into the pipeline.
On the other hand, assuming paraphrasing \textit{impedes} detection, our proposal identifies the most effective attack (and why) so disinformation researchers can prepare a defense for it.
Finally, our research will help determine the best detector for differentiating fake news produced by LLMs.
Understanding which detector provides the most efficacy in this context is essential for devising robust defense mechanisms against misinformation.
 
\boldify{Research questions:}

\begin{itemize}[nosep, leftmargin=27pt]
    \item[\textbf{RQ1}] How do various fake news detectors perform on human-written fake news versus paraphrased versions of the original content?
    \item[\textbf{RQ2}] Which fake news detection models are more robust to paraphrasing?% and why?
    \item[\textbf{RQ3}] Which language models produce paraphrased content that is most difficult or easy to detect as fake news?% and why?   
    \item [\textbf{RQ4}] Which generator provides paraphrased text with high \fbert{}?
    \item[\textbf{RQ5}] What insights can explainability provide about the patterns found in detecting both human-written and paraphrased fake news?
\end{itemize}

%\clearpage
\section{Background
\draftStatus{APPROVED}}

\begin{figure*}
  \includegraphics[width=\linewidth]{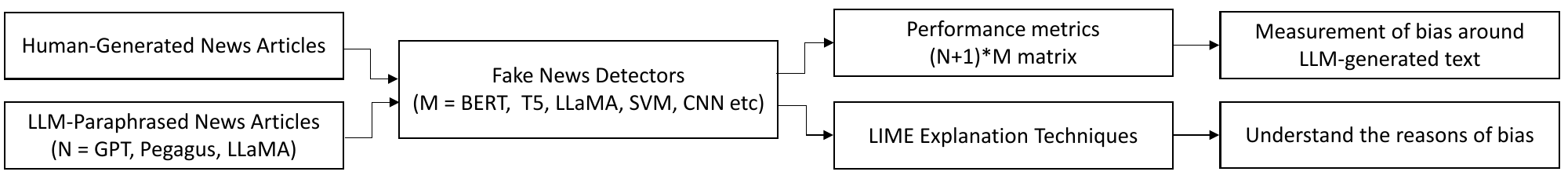}
  \caption{Methodology to assess the efficacy of fake news detectors}
  \label{methodology}
\end{figure*}

\boldify{Fake News Detection: (JED: I think we already covered this enough in the introduction}

\boldify{traditional paraphrasing stretegies}

Researchers have long used traditional techniques to generate paraphrased text, such as manual rules~\cite{mckeown1983paraphrasing} or lexical substitutions~\cite{hassan2007unt}. 
Deep neural networks have been prevalent in generating paraphrased text in the last decade.
\cite{li2017paraphrase} combine a generator and evaluator to design a reinforcement learning-based paraphraser. 
There, the \textit{evaluator} provides the rewards, which then fine-tune the \textit{generator}. 
\cite{gupta2018deep} incorporate an LSTM model with a variational autoencoder (VAE) to generate multiple paraphrased texts for a given sentence. 
A similar kind of VAE model generates paraphrased sentences~\cite{hegde2020unsupervised} without using bilingual data.
\cite{prakash2016neural} utilizes a network of four-layer stacked LSTM and residual connections like the ResNet~\cite{he2016deep} model for paraphrase generation.
\cite{fu2019paraphrase} proposed a similar network with a latent bag of words.
To mitigate the slow training issue in sequence-to-sequence models, \cite{yu2018qanet} proposed a novel approach for paraphrase generation that consists of exclusive convolution for local interactions and self-attention for global interactions.
Researchers also implemented different techniques to generate controlled paraphrased text using an additional set of position embeddings~\cite{goyal2020neural}, decomposition mechanism~\cite{li2019decomposable} or multilayer LSTM~\cite{iyyer2018adversarial}.

\boldify{Pre-trained models in paraphrasing}

Pre-trained models are getting more attention recently in different down-streaming tasks, especially text generation.
Those models are now capable of generating high-quality context~\cite{yu2022generate}. 
\cite{witteveen2019paraphrasing} fine-tune a GPT2 model to generate paraphrased examples. 
A similar work~\cite{hegde2020unsupervised} uses a GPT2 to generate paraphrased text in an unsupervised way without using any labeled data and compares that with other supervised and unsupervised techniques.
Researchers fine-tuned the GPT3 and T5 models to generate paraphrased text and then deployed seven plagiarism detection techniques to detect machine-generated paraphrased text~\cite{wahle2022large}.
\cite{yadav2024pag} generates multiple paraphrases by a Llama model for intent classification.

\boldify{Paraphrasing in data augmentation and dataset creation}

Paraphrasing text for data augmentation is also very popular, especially when there is task-specific data scarcity. 
Researchers used different language models~\cite{jiao2019tinybert,vogel2022investigating}, deep learning models~\cite{kumar2019submodular, hegde2020unsupervised,mi2022improving} to generate paraphrases for data augmentation.
It is an effective technique to implement over-sampling for unbalanced datasets~\cite{patil2022t5w}.
Paraphrasing also increases the accuracy of the model~\cite{kumar2019submodular,hegde2020unsupervised}.
The creation of datasets is another application of paraphrasing.
Researchers generate paraphrase datasets using back-translation~\cite{wieting2017paranmt}, heuristic techniques~\cite{dolan2005automatically}, bilingual pivoting method~\cite{ganitkevitch2013ppdb}, intra-paper and inter-paper method~\cite{dong2021parasci}, a combination of name entities extraction and Jaccard distance metrics~\cite{xu2013gathering}, etc.
%intra-paper:Authors usually write down the same information with transformed expressions repeatedly in different parts of the paper to emphasize critical information or echo back and forth. This kind of feature is the premise of our intra-paper extraction methods.

\boldify{Evaluation techniques by other papers: (CUT CANDIDATE)}

Researchers implemented different automated metrics for evaluating paraphrases, such as BLEU~\cite{li2019decomposable,kumar2019submodular,hegde2020unsupervised,fu2019paraphrase,miao2019cgmh}, METEOR~\cite{gupta2018deep,kumar2019submodular,hegde2020unsupervised}, TER~\cite{gupta2018deep,kumar2019submodular,hegde2020unsupervised}, ROUGE~\cite{goyal2020neural,li2019decomposable,fu2019paraphrase}.
Some authors also appealed to human evaluators along with automatic techniques~\cite{dolan2005automatically,iyyer2018adversarial,goyal2020neural}.

\boldify{Explainability techniques}

Different explainability techniques are available to explain the models to non-technical end users. 
\cite{jin2021euca} proposes the End-User-Centered explainable AI framework EUCA to aid in the end-user-centered XAI design and implementation process.
Those authors describe four categories of ``explanatory forms'': rules, examples,  features, and supplementary information.
Feature attribution-based methods are the most popular ones. 
Researchers use local explanation-based tool LIME~\cite{ribeiro2016should} for local explanation~\cite{szczepanskinew}.
Another popular tool is SHAP~\cite{lundberg2017unified}, which uses local and global explanability~\cite{ayoub2021combat,reis2019explainable}.
Some other feature attribution-based explanation techniques are Integrated Gradients~\cite{sundararajan2017axiomatic} and DeepLIFT~\cite{shrikumar2017learning}.
In prior works, researchers used other techniques, such as causal frameworks~\cite{alvarez2017causal}, attention score-based methods~\cite{neely2022song}, saliency visualizations~\cite{ghaeini2018interpreting}, and Layer-wise Relevance Propagation (LRP)~\cite{gholizadeh2021model} to explain the output of models.

%\clearpage
\section{Methodology
\draftStatus{APPROVED}
}

Figure~\ref{methodology} shows an overview of our methodology.

%\begin{landscape}
\begin{table*}
%\hspace{-85pt}
\footnotesize
\centering
\setlength{\tabcolsep}{4.5pt} % Default value: 6pt
\renewcommand{\arraystretch}{1} % Default value: 1
%\framebox[\linewidth]{For Measuring!}
\begin{tabular}{@{}l| llll| llll| llll| llll @{}}
& \multicolumn{4}{c|}{Human-written} & \multicolumn{4}{c|}{GPT-generated} & \multicolumn{4}{c|}{Llama-generated} & \multicolumn{4}{c}{Pegasus-generated} \\
\hline
%\textbf{\#} &
\textbf{Pipeline} &

\textbf{Acc} &
\textbf{F1} &
\textbf{Pre} &
\textbf{Rec} &

\textbf{Acc} &
\textbf{F1} &
\textbf{Pre} &
\textbf{Rec} &

\textbf{Acc} &
\textbf{F1} &
\textbf{Pre} &
\textbf{Rec} &

\textbf{Acc} &
\textbf{F1} &
\textbf{Pre} &
\textbf{Rec} 
\\\hline

%1  &
BERT                 
% Human-written
& .930             
& .930       
& .930              
& .930           
% GPT-generated
& \textbf{.922}             
& \textbf{.922}      
& \textbf{.922}              
& \textbf{.922}           
% Llama-generated
& .902             
& .902       
& .902              
& .902           
% Pegasus-generated
& .877             
& .877       
& .877              
& .877           
\\

%2 &
T5                   
% Human-written
& .930             
& .932       
& .940              
& .930           
% GPT-generated
& .899            
& .899       
& .901              
& .899           
% Llama-generated
& .904             
& .904       
& .904              
& .904       
% Pegasus-generated
& .868             
& .868       
& .871              
& .868           
\\

%3  &
Llama             
% Human-written
& \textbf{.939}            
& \textbf{.939}       
& \textbf{.940}            
& \textbf{.939}       
% GPT-generated
& .918             
& .918       
& .918              
& .918         
% Llama-generated
& \textbf{.927}             
& \textbf{.927}      
& \textbf{.927}              
& \textbf{.927}           
% Pegasus-generated
& \textbf{.879}             
& \textbf{.879}       
& \textbf{.879}              
& \textbf{.879}           
\\\hline

%4  &
CNN                  
% Human-written
& .920             
& .920       
& .920              
& .920           
% GPT-generated
& .903             
& .903       
& .903              
& .903           
% Llama-generated
& .887             
& .887       
& .887              
& .887           
% Pegasus-generated
& .852             
& .852       
& .852              
& .852           
\\

%5  &
LSTM                 
% Human-written
& .924             
& .924       
& .924              
& .924           
% GPT-generated
& .906             
& .906       
& .906              
& .906           
% Llama-generated
& .895             
& .895       
& .895              
& .895           
% Pegasus-generated
& .868             
& .868       
& .868              
& .868           
\\\hline

%6  &
SVM-cv               
% Human-written
& .914             
& .914       
& .914              
& .914           
% GPT-generated
& .891             
& .891       
& .891              
& .891           
% Llama-generated
& .880            
& .880       
& .880              
& .880           
% Pegasus-generated
& .858             
& .858       
& .859              
& .858          
\\

%7  &
SVM-tfidf            
% Human-written
& .921             
& .921       
& .921              
& .921           
% GPT-generated
& .908             
& .908       
& .908              
& .908           
% Llama-generated
& .896             
& .896       
& .896              
& .896           
% Pegasus-generated
& .864             
& .864       
& .864              
& .864           
\\

%8  &
SVM-wv               
% Human-written
& .874             
& .874       
& .874             
& .874           
% GPT-generated
& .866             
& .866       
& .866              
& .866           
% Llama-generated
& .854             
& .854       
& .854              
& .854           
% Pegasus-generated
& .840             
& .840       
& .841              
& .840           
\\\hline

%9 &
LR-cv                
% Human-written
& .921             
& .921       
& .921              
& .921           
% GPT-generated
& .902             
& .902       
& .903              
& .902           
% Llama-generated
& .893             
& .893       
& .893              
& .893           
% Pegasus-generated
& .866             
& .866       
& .866              
& .866           
\\

%10 &
LR-tfidf             
% Human-written
& .913             
& .913       
& .914              
& .913           
% GPT-generated
& .899             
& .899       
& .900              
& .899          
% Llama-generated
& .890             
& .890       
& .890             
& .890           
% Pegasus-generated
& .863             
& .863       
& .863              
& .863           
\\

%11&
LR-wv               
% Human-written
& .868             
& .868       
& .868              
& .868           
% GPT-generated
& .860             
& .860       
& .860             
& .860           
% Llama-generated
& .852             
& .852       
& .852              
& .852           
% Pegasus-generated
& .837             
& .837       
& .838              
& .837           
\\\hline

%12 &
RF-cv    
% Human-written
& .900           
& .900       
& .900              
& .900           
% GPT-generated
& .886             
& .886       
& .886              
& .886          
% Llama-generated
& .877            
& .877       
& .877              
& .877           
% Pegasus-generated
& .852             
& .852       
& .855              
& .852           
\\

%13 &
RF-tfidf  
% Human-written
& .899           
& .899      
& .900          
& .899         
% GPT-generated
& .885            
& .885       
& .885          
& .885           
% Llama-generated
& .871            
& .871       
& .871           
& .871      
% Pegasus-generated
& .851             
& .851       
& .852            
& .851           
\\

%14 &
RF-wv   
% Human-written
& .868            
& .868      
& .871         
& .868         
% GPT-generated
& .850          
& .850       
& .852          
& .850          
% Llama-generated
& .835             
& .835       
& .837            
& .835          
% Pegasus-generated
& .825          
& .825      
& .828             
& .825     
\\\hline

%15 &
DT-cv                
% Human-written
& .856             
& .855       
& .855              
& .856           
% GPT-generated
& .807             
& .807       
& .807              
& .807           
% Llama-generated
& .793             
& .793       
& .793             
& .793           
% Pegasus-generated
& .804             
& .804       
& .804              
& .804           
\\

%16 &
DT-tfidf  
% Human-written
& .846             
& .846       
& .846              
& .846           
% GPT-generated
& .807             
& .806       
& .808              
& .807           
% Llama-generated
& .785             
& .785       
& .785              
& .785           
% Pegasus-generated
& .786             
& .786       
& .786              
& .786           
\\

%17 &
DT-wv  
% Human-written
& .770             
& .770       
& .770              
& .770           
% GPT-generated
& .742             
& .742       
& .742              
& .742           
% Llama-generated
& .735             
& .735       
& .737              
& .735           
% Pegasus-generated
& .721             
& .722       
& .723              
& .721           

%%%%%%%%%%%%%%%%%%%%%%%%%%%%%%%%%%%%%%%%%%

\\\hline

\end{tabular}
\caption{Classification performance for human-written vs LLM paraphrased Covid-19 dataset}
\label{table_covid-19}
\end{table*}
%\end{landscape}

%%%%%%%%%%%%%%%%%%%%%%%%%%%%%%%%%%%%%%%%%%

%\subsection{Datasets}

We used two publicly available datasets to assess the performance of fake news detectors.
The first dataset~\cite{patwa2020fighting} is on \covid{} misinformation, containing only two classes: \texttt{Real} and \texttt{Fake}. 
This dataset has 5524 real news and 5030 fake news, making it quite balanced.
The second dataset comes from POLITIFACT.com and is called LIAR~\cite{wang2017liar}.
It consists of 12.8K manually classified statements made by politicians in various contexts.
% There are a total of 14 fields in the dataset, such as ``the speaker,'' ``the subject(s),'' ``the ID of the statement,'' ``the label'' and ``the statement.''
However, this dataset differs from the \covid{} dataset because it has six labels: true—16.37\%, mostly true—19.15\%, half true—20.68\%, barely true—16.22\%, false—19.50\%, and pants-fire— 8.09\%.
The creators have pre-split the dataset into train, test, and validation.
We curated and preprocessed data with the  NLTK~\cite{loper2002nltk} library to prepare the data for input to the detectors. 

\subsection{Classifiers and Paraphrasers}

%After that, we investigated the ability of fake news detectors to identify human-written fake news. 
We considered logistic regression, decision tree, random forest, and support vector machine as supervised models, CNN and LSTM as deep learning models, and BERT~\cite{devlin2018bert}, T5~\cite{raffel2020exploring} and Llama~\cite{touvron2023llama} as pre-trained language models.
We selected those models due to their effectiveness and popularity in  text classification tasks~\cite{gasparetto2022survey}.

In the next step, we used three techniques, each from an LLM family, to paraphrase both the fake and real news.
The first paraphraser is called PEGASUS~\cite{zhang2020pegasus}, a transformer-based model.
% The second model is the Parrot~\cite{prithivida2021parrot} from HuggingFace and trained on a T5 model. 
For the other two methods, we generated paraphrases with the GPT and Llama API.

\subsection{Implementation Details}

\boldify{Rough structure is one paragraph per model}

We performed our experiment on a computer with NVIDIA RTX A4500 (20GB) GPU, consuming a total of $\approx$20 hours.
We implemented the supervised learning algorithms from the sklearn python library~\cite{pedregosa2011scikit}.

We built the CNN models in TensorFlow.
The input layer consisted of 1024 units with ReLU activation, followed by hidden layers with 512 and 256 units also activating ReLU, a dropout layer with a 0.2 rate, and an output layer with units equivalent to the number of classes and sigmoid activation for multi-class classification.
We employed a batch size of 32, 10 training epochs, and an embedding size of 300 in building the CNN model.

Our TensorFlow-based LSTM classifier consisted of an LSTM layer with 100 units, a dense output layer with sigmoid activation, an LSTM layer with the pre-trained embeddings, and a spatial dropout1D layer.

We trained the PyTorch-based $BERT_{base}$ model on the GPU for ten epochs with: learning rate at 1e-5, Adam epsilon at 1e-8, and tokenized text sequences capped at 300 characters.

We adopted the T5 classifier and parameters from a GitHub repository~\cite{Taghizadeh2023}.

\subsection{Evaluation Techniques}

To evaluate the performance of the models, we adopted the typical metrics (i.e., accuracy, F1 score, precision, and recall).
To determine which detectors perform best for a given task, we relied on the macro-F1 score because it balances the importance of precision and recall, especially for an imbalanced dataset. 
The LIAR dataset is quite imbalanced, and the F1-score is the better evaluation metric to access the classification result.
We measured the performance of the same set of detectors on both the original and paraphrased texts.

We evaluated the paraphraser's quality with the open-source contextual embedding-based technique BERTSCORE, which showed strong performance in adversarial paraphrase detection~\cite{zhang2019bertscore}.
Our specific metric is the F1 value of BERTSCORE, which we denote as \fbert{}.
Calculating similarity with \fbert{} takes two arguments, which are, in our case, human-written fake news and the LLM-paraphrased output.
This \fbert{} is the harmonic mean of $P_{BERT}$ and $R_{BERT}$.
$P_{BERT}$ measures how much the reference sentence captures the meaning of the candidate sentence by averaging the maximum cosine similarity between each token in the candidate and the reference.
$R_{BERT}$ measures how much the candidate sentence captures the meaning of the reference sentence using the same technique.

Finally, we explored LIME explanations to discover reasons for getting different classification results between human-written and paraphrased text.
Based on those observations, we then applied a sentiment analyzer~\cite{yuan10distilbert} to each tuple of human-written and three LLM-paraphrased outputs.

%\clearpage
\section{Results}

Our Supplemental Materials contain two files which have all of the data we used: \textit{``Original text''}, three \textit{``<MODEL> paraphrased''}, each with accompanying paraphrase and sentiment scores.

%\begin{landscape}
\begin{table*}
%\hspace{-85pt}
\footnotesize
\centering
\setlength{\tabcolsep}{4.5pt} % Default value: 6pt
\renewcommand{\arraystretch}{1} % Default value: 1
%\framebox[\linewidth]{For Measuring!}
\begin{tabular}{@{}l| llll| llll| llll| llll @{}}
& \multicolumn{4}{c|}{Human-written} & \multicolumn{4}{c|}{GPT-generated} & \multicolumn{4}{c|}{Llama-generated} & \multicolumn{4}{c}{Pegasus-generated} \\
\hline
%\textbf{\#} &
\textbf{Pipeline} &

\textbf{Acc} &
\textbf{F1} &
\textbf{Pre} &
\textbf{Rec} &

\textbf{Acc} &
\textbf{F1} &
\textbf{Pre} &
\textbf{Rec} &

\textbf{Acc} &
\textbf{F1} &
\textbf{Pre} &
\textbf{Rec} &

\textbf{Acc} &
\textbf{F1} &
\textbf{Pre} &
\textbf{Rec} 
\\\hline

%1 &
BERT 
% Human-written
& .251             
& .232       
& .238              
& .251           
% GPT-generated
& .266             
& \textbf{.251}       
& .272              
& .266          
% Llama-generated
& .256             
& .243       
& .276              
& .256           
% Pegasus-generated
& .256             
& .238       
& .270              
& .256\\

%2 &
T5     
% Human-written
& \textbf{.274 }            
& .236       
& \textbf{.312}              
& \textbf{.274}          
% GPT-generated
& \textbf{.277}             
& .241       
& \textbf{.303}              
& \textbf{.277}           
% Llama-generated
& \textbf{.265}             
& \textbf{.262}       
& .264              
& .265           
% Pegasus-generated
& \textbf{.272}             
& \textbf{.270}      
& \textbf{.275}              
& \textbf{.272}\\

%3 &
Llama 
% Human-written
& .253             
& .201       
& .273              
& .253           
% GPT-generated
& .269             
& .236       
& .264              
& .264           
% Llama-generated
& .258             
& .194       
& .204              
& .258           
% Pegasus-generated
& .217             
& .154       
& .259              
& .217
\\\hline

%4 &
CNN     
% Human-written
& .231             
& .224       
& .224              
& .231           
% GPT-generated
& .221             
& .220       
& .219              
& .221           
% Llama-generated
& .239             
& .238      
& .237              
& .239          
% Pegasus-generated
& .213             
& .210      
& .213              
& .213\\

%5 &
LSTM     
% Human-written
& .255             
& \textbf{.258}       
& .255              
& .238           
% GPT-generated
& .234            
& .229       
& .228              
& .234           
% Llama-generated
& .251             
& .247       
& .253              
& .251           
% Pegasus-generated
& .212             
& .192       
& .214              
& .212
\\\hline

%6 &
SVM-cv               
% Human-written
& .227             
& .221       
& .221              
& .223           
% GPT-generated
& .213             
& .211       
& .211              
& .213           
% Llama-generated
& .227             
& .226       
& .228              
& .227           
% Pegasus-generated
& .218             
& .217       
& .218              
& .218\\

%7 &
SVM-tfidf            
% Human-written
& .238             
& .230       
& .229              
& .231           
% GPT-generated
& .242             
& .235       
& .245              
& .242           
% Llama-generated
& .259             
& .254       
& .260              
& .259           
% Pegasus-generated
& .226             
& .218       
& .221              
& .226\\

%8 &
SVM-wv               
% Human-written
& .214             
& .165       
& .228              
& .198           
% GPT-generated
& .248             
& .232       
& .245              
& .248           
% Llama-generated
& .243             
& .230       
& .245              
& .243           
% Pegasus-generated
& .235             
& .221       
& .229              
& .235
\\\hline

%9 &
LR-cv                
% Human-written
& .239             
& .227       
& .230              
& .226           
% GPT-generated
& .240             
& .238       
& .240              
& .240           
% Llama-generated
& .236             
& .233       
& .233              
& .236           
% Pegasus-generated
& .220             
& .217       
& .216              
& .220\\

%10 &
LR-tfidf             
% Human-written
& .238             
& .213       
& .233              
& .215           
% GPT-generated
& .250             
& .242       
& .252              
& .250           
% Llama-generated
& .244             
& .239       
& .250              
& .244           
% Pegasus-generated
& .228             
& .220       
& .219              
& .228 \\

%11 &
LR-wv                
% Human-written
& .246             
& .220       
& .232              
& .223           
% GPT-generated
& .250             
& .244       
& .249              
& .250           
% Llama-generated
& .242             
& .236       
& .244              
& .242           
% Pegasus-generated
& .245             
& .237       
& .239              
& .245
\\\hline

%12 &
RF-cv                
% Human-written
& .250            
& .222       
& .261              
& .227           
% GPT-generated
& .253             
& .235       
& .249              
& .253           
% Llama-generated
& .268             
& .256       
& \textbf{.277}              
& .268           
% Pegasus-generated
& .227             
& .220       
& .223              
& .227\\

%13 &
RF-tfidf   
% Human-written
& .261             
& .227       
& .257              
& .234           
% GPT-generated
& .252             
& .241       
& .253              
& .252           
% Llama-generated
& .272             
& .263       
& \textbf{.277}             
& \textbf{.272}          
% Pegasus-generated
& .224             
& .215       
& .227              
& .224\\

%14 &
RF-wv  
% Human-written
& .231             
& .204       
& .267              
& .207           
% GPT-generated
& .253             
& .235       
& .249              
& .253           
% Llama-generated
& .226             
& .213       
& .234              
& .226           
% Pegasus-generated
& .227             
& .214       
& .225              
& .227
\\\hline

%15 &
DT-cv          
% Human-written
& .233             
& .222       
& .229              
& .222           
% GPT-generated
& .222             
& .219       
& .219              
& .222           
% Llama-generated
& .234             
& .233       
& .234              
& .234           
% Pegasus-generated
& .210             
& .209       
& .209              
& .210\\

%16 &
DT-tfidf             
% Human-written
& .201             
& .192       
& .193              
& .193           
% GPT-generated
& .204             
& .199       
& .197              
& .204           
% Llama-generated
& .199             
& .197       
& .197              
& .199           
% Pegasus-generated
& .209             
& .208       
& .207              
& .209\\

%17 &
DT-wv      
% Human-written
& .180             
& .172       
& .172              
& .172           
% GPT-generated
& .179             
& .179       
& .180              
& .179           
% Llama-generated
& .195             
& .195       
& .196              
& .195           
% Pegasus-generated
& .192             
& .192       
& .193              
& .192

%%%%%%%%%%%%%%%%%%%%%%%%%%%%%%%%%%%%%%%%%%

\\\hline

\end{tabular}
\caption{Classification performance for human-written vs LLM paraphrased LIAR-6 dataset}
\label{table_liar_6}
\end{table*}
%\end{landscape}

%%%%%%%%%%%%%%%%%%%%%%%%%%%%%%%%%%%%%%%%%%

\subsection{RQ1 - Human-writing vs Paraphrase
\draftStatus{APPROVED}}

\boldify{Remind reader of the RQ, then answer it DIRECTLY}

In RQ1, we find out how the detectors perform on human-written fake news and LLM-paraphrased fake news articles.
The results show that the \textbf{detectors struggle to detect LLM-generated fake news more than human-written fake news}.
Further, while the F1 scores were low for LLM-paraphrased fake news, the \textbf{Pegasus-paraphrased fake news is the most challenging to detect}.

\boldify{Describe the key analysis points for the "easy" dataset}

Table~\ref{table_covid-19} demonstrates the accuracy, F1 score, precision, and recall values of all the fake news detectors on the \covid{} dataset, and Figure~\ref{F1_comparison} (Top) compares their F1 score. 
All 17 detectors achieve a high F1 score, with human-written news articles being easiest to detect. 
Additionally, this dataset exhibits consistent results, with human-written fake news being the easiest for all the detectors.

\boldify{Describe the key analysis points for the "hard" dataset}

Table~\ref{table_liar_6} shows the performance of fake news detectors on human-written vs LLM-paraphrased news articles on the LIAR dataset. 
Figure~\ref{F1_comparison} (Bottom) compares F1 scores among all the detectors.
The Figure and Table both show that no source was easier or harder to detect consistently.
Specifically, encoder-decoder models (e.g., BERT, T5, Llama) yield low F1 score in detecting human-written fake news articles, when compared to GPT or Llama-paraphrased fake news. 
Both deep learning methods (CNN and LSTM) attain a low F1 score in classifying Pegasus and GPT-paraphrased news articles. 
On the other hand, supervised learning models such as SVM, logistic regressions, and random forests, regardless of the features (TF-IDF, Countvectorizer, Word Embeddings), show a high F1 score in identifying GPT and Llama-paraphrased texts, which indicates their struggle in detecting human-written and Pegasus-paraphrased fake news. 

%\clearpage
\subsection{RQ2 - Detector Efficacy
\draftStatus{APPROVED}}

\boldify{Remind them of the RQ, and answer it directly}

In RQ2, we determine which fake news detector is more robust in detecting fake news generated by LLM.
Here, we have a mixed result: \textbf{For the \covid{} dataset, the LLM-based models are superior, but the LIAR dataset is less clear}.

\boldify{Add any nuances desired from the easy dataset}

\begin{figure}
  \includegraphics[width=\linewidth]{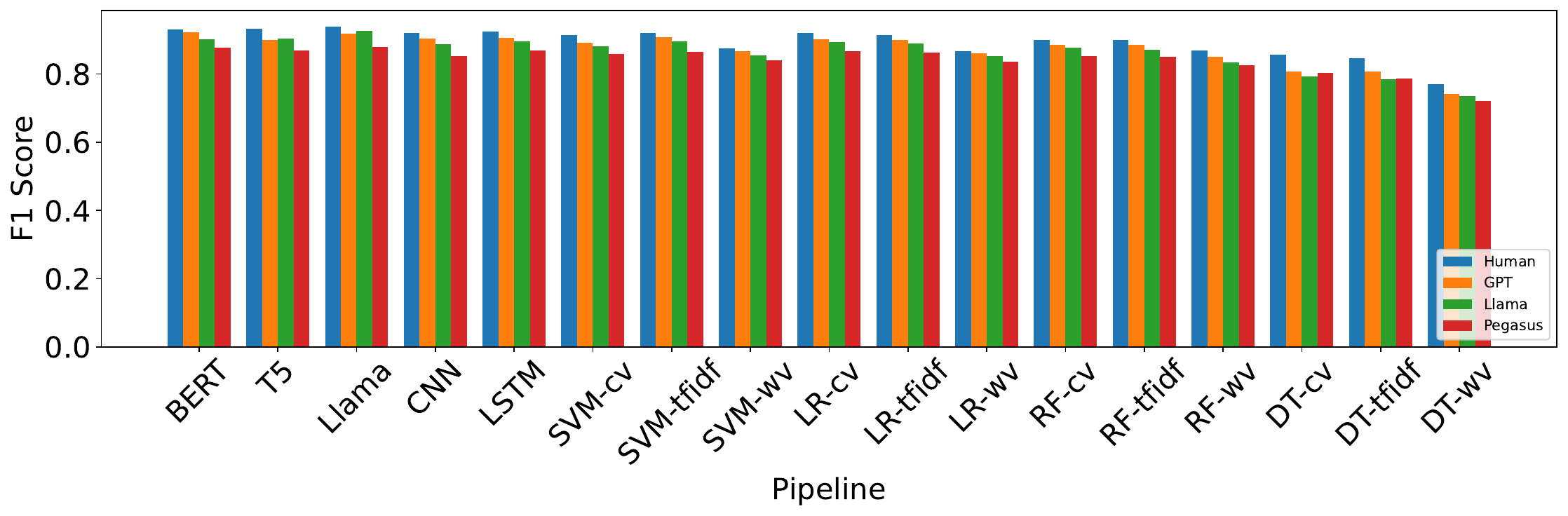}

  \includegraphics[width=\linewidth]{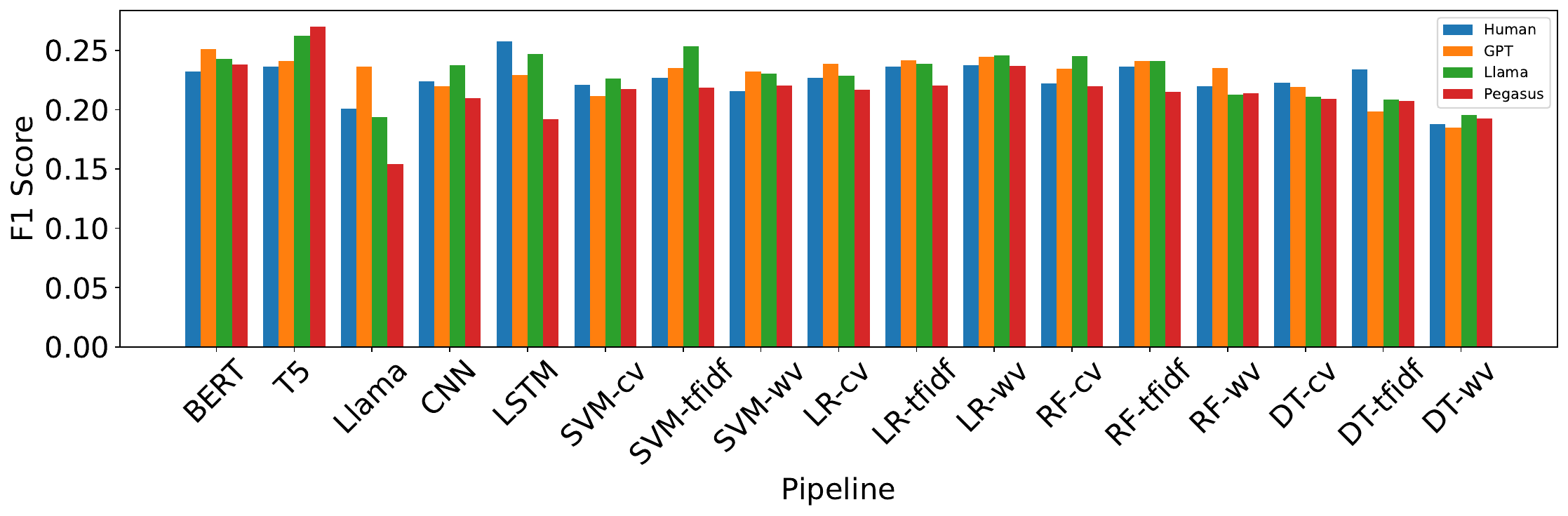}
  \caption{(\textbf{Top}): Performance of fake news detectors on human-written and LLM-paraphrased text on \covid{} dataset.
  (\textbf{Bottom}): Same, but on LIAR dataset}
  \label{F1_comparison}
\end{figure}

For the \covid{} dataset, LLM-based detectors (BERT, Llama, and T5) and deep learning-based models (LSTM and CNN) all perform well (Figure~\ref{F1_comparison} (Top) and Table~\ref{table_covid-19}). 
Supervised learning models, such as SVM with TF-IDF features, also show moderate performance. 
All LLM-based detectors are excellent in detecting all three LLM-paraphrased fake news, followed by deep learning models and SVM with TF-IDF features.

\boldify{Add any nuances desired from the hard dataset}

For the LIAR dataset, all detectors display inconsistent results (Figure~\ref{F1_comparison} (Bottom) and Table~\ref{table_liar_6}). 
LSTM is better at detecting human-written fake news than LLM-based detectors.
For GPT-paraphrased fake news, the BERT model is the best, followed by logistic regression with TF-IDF and word embeddings, and then T5.
T5 achieves the best F1 scores, followed by SVM and LSTM in Llama-paraphrased text. 
Even the fine-tuned Llama model cannot produce a good F1 score in the llama-paraphrased text.
It indicates that even LLM-based detectors struggle to detect self-generated text, as mentioned in~\cite{jiang2023disinformation}

%\clearpage
\subsection{RQ3 - Paraphraser Detectability
\draftStatus{APPROVED}}

In RQ3, we find out which LLM-paraphrased fake news was hard/easy to detect.
In general, for both datasets, \textbf{fake news detectors struggle with the fake news from Pegasus more than Llama, which was more of a struggle than GPT}.

Especially in the \covid{} dataset (Figure~\ref{F1_comparison} (Top) and Table~\ref{table_covid-19}), the F1 scores of all detectors for the Pegasus-paraphrased dataset are the worst.
In the LIAR dataset (Figure~\ref{F1_comparison} (Bottom) and Table~\ref{table_liar_6}), the F1 scores for the 11 detectors (among 17) are worst for Pegasus-generated fake news.
%Llama is the next paraphraser that most detectors have trouble identifying. 

15 of the 17 detectors exhibit the second lowest F1 score for the Llama-paraphrased text for the \covid{} data set. 
On the other hand, GPT-paraphrased fake news is easy to detect for most of the classifiers for both datasets, but not as easy as the original human-written text.
%\clearpage
\subsection{RQ4 - Paraphraser BERTSCORE
\draftStatus{APPROVED}
}

\begin{figure}
    \centering

    \includegraphics[width=\linewidth]{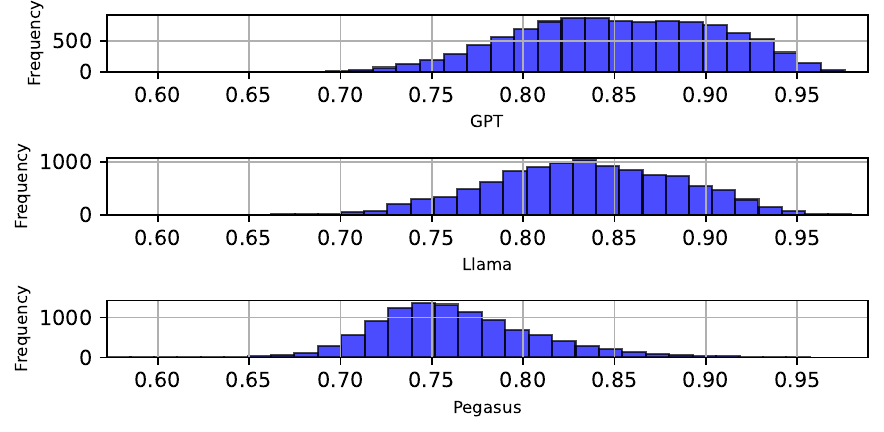}
    
    \caption{Distribution of $F_{BERT}$ score for all paraphrasers on COVID-19 dataset.
    Higher is better.}
    \label{covid_bertscore}
\end{figure}

\begin{figure}
    \centering

    \includegraphics[width=\linewidth]{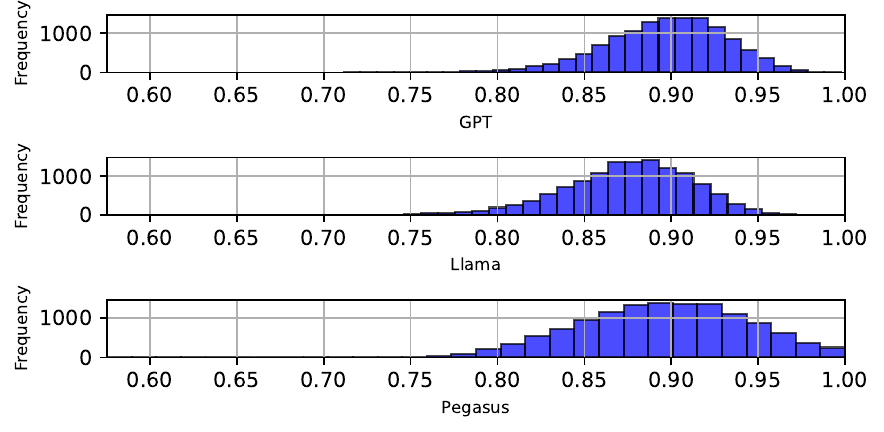}
    
    \caption{Distribution of $F_{BERT}$ score for all paraphrasers on LIAR dataset}
    \label{liar_bertscore}
\end{figure}

Our RQ4 was to find out which paraphraser generates the highest quality paraphrases, as measured by the \fbert{}~\cite{zhang2019bertscore}.
In general, \textbf{GPT emerges as the most reliable tool for maintaining high semantic similarity} in \covid{} and LIAR data sets.
Figures~\ref{covid_bertscore} and \ref{liar_bertscore} illustrate the semantic similarity score distributions.

We also calculated Hedge’s g to measure the effect sizes on \fbert{} between treatments, here using different paraphrasers.
For the \covid{} dataset, we find a small effect size between GPT and Llama (Hedge’s g, 0.34), which indicates low difference in the semantic similarity between their paraphrased text outputs.
In contrast, we find very large effect sizes between GPT and Pegasus (Hedge's g, 1.78) and between Llama and Pegasus (Hedge's g, 1.47), which indicates that GPT and Llama produce paraphrases with practically significantly higher semantic similarity than Pegasus.
For the LIAR dataset, we find negligible effects between Llama and  paraphrasers (Hedge's g, $|g| < .06$), but medium effect size between GPT and Llama (Hedge's g, 0.60), which substantiates our earlier observation about the superior \fbert{}s that GPT paraphrases possess.

%\clearpage
\subsection{RQ5 - Explainability
\draftStatus{APPROVED}}
\label{sectionRQ5}

\begin{figure*}
    \begin{tabular}{@{} r | l @{}}
  \includegraphics[width=.48\linewidth]{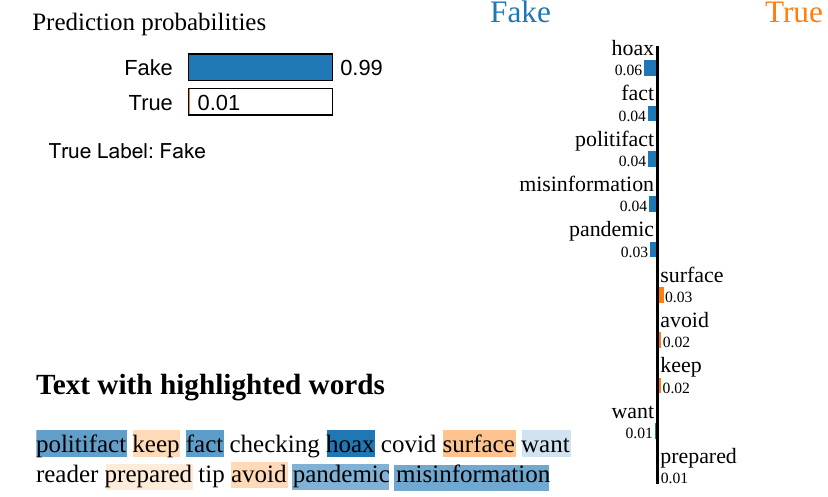}
  &
  \includegraphics[width=.48\linewidth]{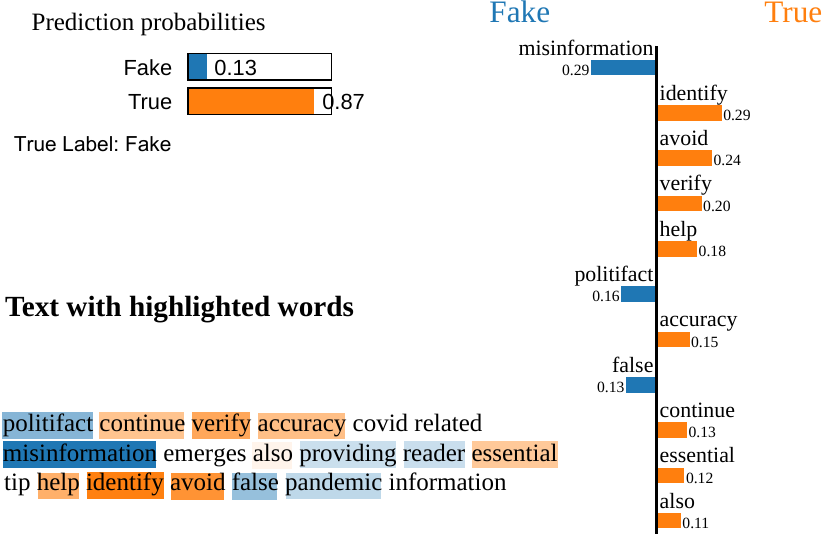}
\\\hline

  \includegraphics[width=.48\linewidth]{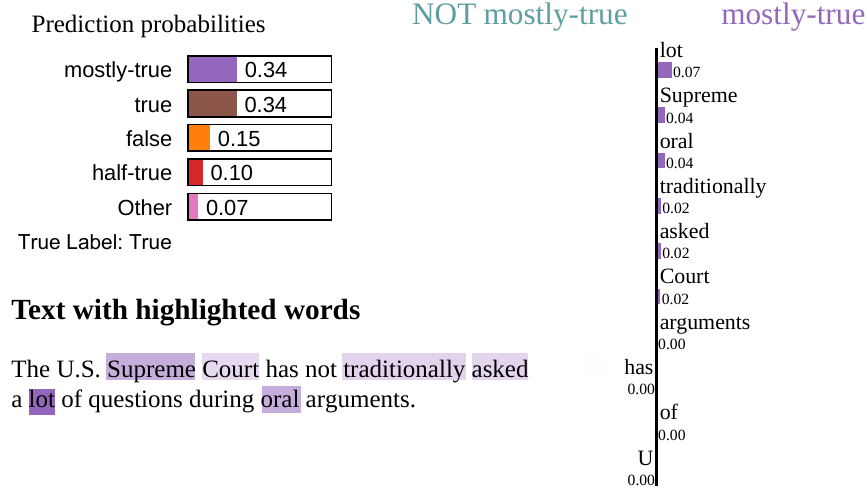}
  &
  \includegraphics[width=.48\linewidth]{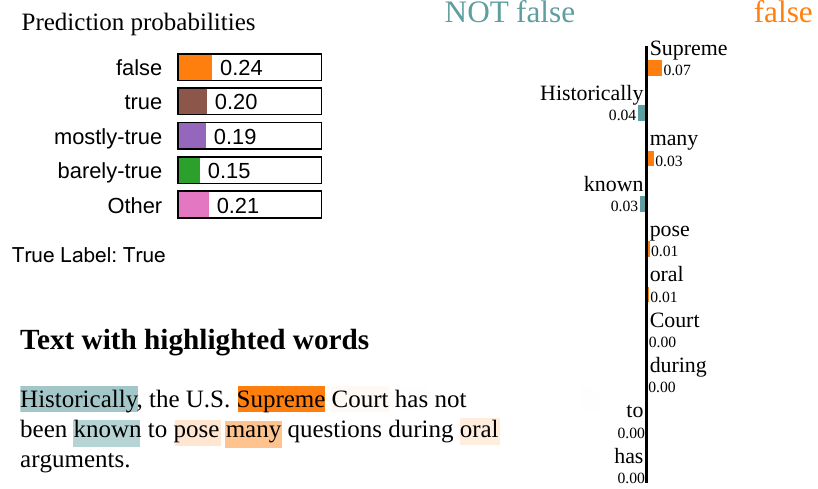}
    \end{tabular}

  \caption{(\textbf{Top Left}): LIME output of the BERT model on human-written news
  (\textbf{Top Right}): LIME output of the BERT model on Llama-paraphrased news
  (\textbf{Bottom Left}): LIME output of the LSTM model on human-written news
  (\textbf{Bottom Right}): LIME output of the LSTM model on GPT-paraphrased news}
  \label{voltron}
\end{figure*}

Table~\ref{table_covid-19} shows that the BERT model performs well in human-written news articles.
However, its performance decreases with Llama-paraphrased news articles. 
To observe the reason, we selected an instance where the BERT model could accurately classify a piece of human-written fake news, but failed to classify the Llama-paraphrased version.

The original sentence (Figure~\ref{voltron} (Top Left)) contains \textit{``hoax''}, \textit{``covid''}, \textit{``misinformation''}, and \textit{``avoid''} pandemic misinformation, which tend to have negative sentiment. 
The LIME explainer also assigns high weight values on those words and BERT correctly classifies the sentence as \texttt{Fake}.

On the other hand, the Llama-paraphrased version (Figure~\ref{voltron} (Top Right)) contains a lot of positive sentiments, such as \textit{``verify accuracy'',} \textit{``essential tip''}, and \textit{``help identity avoid false pandemic information''} and mispredicted the sentence as \texttt{True}.
Next, we measured the shift in sentiment between the original and paraphrase with Amazon Comprehend, a cloud platform for sentiment analysis.
In this case, the human-written text was judged mostly negative 
(0.58 negative and 0.01 positive sentiment scores), while llama-paraphrased text was slightly positive (%0.66 neutral, 
0.21 positive and 
0.08 negative).
Thus, a sentiment shift might be the reason for this misclassification of the BERT model.

For the LIAR dataset, we see a similar performance degradation---the LSTM model achieves an F1 score of 0.258 on human-written text, but then reduces to 0.229 on GPT-paraphrased text. 
Again, we selected an instance where the LSTM accurately classified the human-written text, but misclassified the GPT-paraphrased text.

Figure~\ref{voltron} (Bottom Left) shows that human-written text contains words like \textit{``Traditionally''} and may have a neutral or factual tone associated with norms and history, which the model interprets as aligning with \texttt{True} statements.

On the other hand, in the paraphrase (Figure~\ref{voltron} (Bottom Right)), \textit{``not known to''} introduces ambiguity.
This framing could trigger the model to infer doubt or unreliability, causing it to classify the statement as \texttt{False}.
This ambiguity might assign a higher weight towards \texttt{False} for the word \textit{``Supreme''} for the GPT-paraphrased text.
However, the same \textit{``Supreme''} word has a higher weight towards \textit{``mostly true''} for the human-written text.

The LIME explanations suggest that a shift in sentiment and/or introduction of ambiguity during paraphrasing might be the reason for making fake news detection hard for the detectors.

After that, we looked at the possibility of the sentiment shift in more detail. 
We used a HuggingFace model for sentiment analysis that takes a sentence and returns a dictionary of sentiment values.
To calculate the sentiment shift, we considered the positive and negative sentiments returned by the sentiment analyzer for both human-written and LLM-paraphrased text. 
We ignored the neutral sentiment in our calculation.
Assuming a three tuple (+, 0, -), we compute a difference of differences between positive sentiment $P$ and negative sentiment $N$, each of which can come from human (e.g., $N_h$) or LLM (e.g., $P_l$):
\begin{equation}
\label{eqnSentimentShift}
S = (P_h - N_h) - (P_l - N_l)
\end{equation}
Note that when $S$ is positive, the paraphrase has more negative sentiment than the human-written text, and vice-versa when $S$ is negative.

\begin{figure}
  \includegraphics[width=\linewidth]{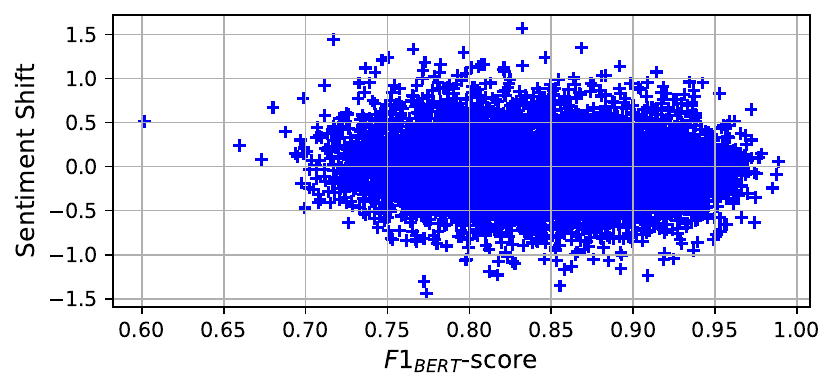}
  \caption{\fbert{} vs sentiment shift (Human-GPT) on \covid{} dataset.
  Here, we plotted only this configuration, as all other configurations have patterns.}
  \label{sentiment_shift}
\end{figure}

Then we created a scatter plot of the values of the sentiment shift against the \fbert{} (Figure~\ref{sentiment_shift}).
We observed that the data followed pattern similar to the figure for all the LLM paraphrasers. 
The figure shows that a substantial amount of text has a higher \fbert{} score, but their sentiment shifted during the paraphrasing process.

Consider that a point with $|S| > 1$ will \textit{definitely} have flipped sentiment as a result of the paraphrase process.
Approximately 0.45\% data points in this figure have $|S| > 1$.
Similarly, a point with $|S| > 0.5$ is \textit{more probable than not} to have flipped sentiment.
Approximately 6.86\% data points have $|S| > 0.5$.
A good paraphrased text should have a high semantic similarity (high \fbert{} in this case) and a low semantic shift ($|S|$).

Oddly, we find human-written and paraphrased versions of that text sometimes have high semantic similarity, but convey different sentiments.
%That provides an important insight into the misclassification of the fake news detectors in the paraphrased text.
%JED: removed this sentence. Why? this sentence is weak and the new last sentence is strong and forms a good segue into the disussion section

%\clearpage
\section{Discussion
\draftStatus{APPROVED}
}

\subsection{Which Quality Measures?}

In RQ3 and RQ4, we introduced two different ways to measure the quality of a paraphrase.
RQ4 simply appealed to the \fbert{} metric, while RQ3 measured the change in detection rates between the paraphrase and the original human-written text.
By the \fbert{} metric, GPT is our best-performing paraphrasing model.
However, when we look at detection F1, we see a contradictory result, namely that one of the other models are better.
Specifically, if we interpret a larger reduction in F1 to be ``better'' (i.e., the paraphrase is more able to conceal the true label of the text), then Pegasus is best.
Meanwhile, if we interpret a smaller reduction in F1 to be ``better'' (i.e., the paraphrase retains as much of the labeling of the original as possible), then Llama is best.
As a result, a number of open questions swirl, such as which measure we should rely upon more and why, as well as how to devise a top-down paraphrase metric that better aligns with bottom-up observations.

\subsection{Changing Sentiment Without Changing Semantics?}

In Section~\ref{sectionRQ5} we saw evidence that a lot of data had large sentiment shifts, but high \fbert{}.
This seems like a rather large problem for two reasons.
First, as we already mentioned, it seems to be introducing confusion into the classification problem.
Second, this combination should not be possible from the metric perspective.
Ultimately, \fbert{} relies on some combination of semantic similarity and syntactic similarity.
One interpretation of our results is that this combination may benefit from more terms, (e.g., sentimental similarity).
Ultimately, our results indicate that there is room for researchers to improve some aspect of semantic similarity measurement.

%\clearpage
\section{Conclusion}

\boldify{Contributions-style conclusion!}

In answering our RQs, we made five contributions.
First, paraphrasing tends to decrease classification accuracy, indicating that LLM laundering can be an effective attack to evade fake news detectors.
Second, we identify which models perform well at which tasks.
Specifically, we found LLM-based models to be the detector most robust to LLM-paraphrased text, Pegasus to be the generator that best evades detection and GPT to be the generator that creates most semantically similar fake news paraphrases, as measured by \fbert{}.
Third, LIME explanations revealed a possible reason for failures, specifically \textit{sentiment shift}.
Fourth, we provide evidence about the prevalence of a shift in sentiment paired with a high semantic similarity.
Finally, we introduce two paraphrased datasets for future researchers to build more robust models and techniques for detecting fake news.

%1. paraphrased text is more likely to slip past a detector (RQ1)
%2. Which models tend to perform well at which tasks (RQ2/3/4)
%3. Explainability results find a possible reason for detection failures (RQ5)
%4. Worrisome observation about efficacy of the metric: shift in sentiment, despite high bertscore are concerning (RQ4/5)
%5. a dataset of our paraphrases (ALL RQs)

%\clearpage
\section{Limitations}

% Limitation is mandatory for this submission. 

% Long Papers

% Long papers must describe substantial, original, completed and unpublished work. Wherever appropriate, concrete evaluation and analysis should be included. Long papers may consist of:

%     up to eight (8) pages of content
%     plus up to one page for limitations (required, see below) and optionally ethical considerations
%     plus unlimited pages of references

% Submissions that exceed the length requirements, or are missing a limitations section, will be desk rejected.

Despite the insights provided by this study, it still has some limitations. 
This study focuses on a limited number of paraphrasing models (GPT, Llama, and Pegasus) and LLM-based detectors (Llama, T5, and BERT).
Similarly, we only covered a limited set of datasets (\covid{} and LIAR).
Further, the LIAR dataset is rather strange---the text instances are short and the leaderboard models all perform poorly (.27\% Accuracy).

While we focus on semantic similarity using BERTScore, we do not examine other dimensions of text quality, such as fluency or coherence.
An empirical study with human evaluators might be a better evaluation technique to identify the best paraphrases.

In this paper, we proposed our own metric for semantic shift (Equation~\ref{eqnSentimentShift}).
While the formula is rather straightforward and makes sense, this is not a validated process.

Though this study finds that a shift in sentiment and introduction of ambiguity might be the reason for the detectors to detect fake news properly, we also need a more comprehensive study to ascertain that claim.

\section{Ethical Considerations}

In this work, we enumerate a potential method to improve evading fake news detectors.
As with much work in security, enumerating an attack always poses the risk that malicious actors deploy the attack.
However, the hope is that the defenders' awareness of the attack counterbalances this concern, since they are able to develop mitigation strategies for that specific attack and avoid being surprised by it.
%\clearpage

\pagestyle{empty}
]\bibliographystyle{unsrt}
\bibliography{00-paper}

\begin{thebibliography}{10}

\bibitem{cao2017joint}
Ziqiang Cao, Chuwei Luo, Wenjie Li, and Sujian Li.
\newblock Joint copying and restricted generation for paraphrase.
\newblock In {\em Proceedings of the AAAI Conference on Artificial Intelligence}, volume~31, 2017.

\bibitem{berant2014semantic}
Jonathan Berant and Percy Liang.
\newblock Semantic parsing via paraphrasing.
\newblock In {\em Proceedings of the 52nd Annual Meeting of the Association for Computational Linguistics (Volume 1: Long Papers)}, pages 1415--1425, 2014.

\bibitem{fader2014open}
Anthony Fader, Luke Zettlemoyer, and Oren Etzioni.
\newblock Open question answering over curated and extracted knowledge bases.
\newblock In {\em Proceedings of the 20th ACM SIGKDD international conference on Knowledge discovery and data mining}, pages 1156--1165, 2014.

\bibitem{yin2015answering}
Pengcheng Yin, Nan Duan, Ben Kao, Junwei Bao, and Ming Zhou.
\newblock Answering questions with complex semantic constraints on open knowledge bases.
\newblock In {\em Proceedings of the 24th ACM international on conference on information and knowledge management}, pages 1301--1310, 2015.

\bibitem{yu2018qanet}
Adams~Wei Yu, David Dohan, Minh-Thang Luong, Rui Zhao, Kai Chen, Mohammad Norouzi, and Quoc~V Le.
\newblock Qanet: Combining local convolution with global self-attention for reading comprehension.
\newblock {\em arXiv preprint arXiv:1804.09541}, 2018.

\bibitem{iyyer2018adversarial}
Mohit Iyyer, John Wieting, Kevin Gimpel, and Luke Zettlemoyer.
\newblock Adversarial example generation with syntactically controlled paraphrase networks.
\newblock {\em arXiv preprint arXiv:1804.06059}, 2018.

\bibitem{khurana2023natural}
Diksha Khurana, Aditya Koli, Kiran Khatter, and Sukhdev Singh.
\newblock Natural language processing: state of the art, current trends and challenges.
\newblock {\em Multimedia tools and applications}, 82(3):3713--3744, 2023.

\bibitem{openAIreport}
Ina Fried.
\newblock Openai report details election interference efforts, hoaxes.
\newblock \url{https://www.axios.com/2024/10/09/openai-election-interference-political-misinformation}, 2024.

\bibitem{brown2020language}
Tom Brown, Benjamin Mann, Nick Ryder, Melanie Subbiah, Jared~D Kaplan, Prafulla Dhariwal, Arvind Neelakantan, Pranav Shyam, Girish Sastry, Amanda Askell, et~al.
\newblock Language models are few-shot learners.
\newblock {\em Advances in neural information processing systems}, 33:1877--1901, 2020.

\bibitem{devlin2018bert}
Jacob Devlin, Ming-Wei Chang, Kenton Lee, and Kristina Toutanova.
\newblock Bert: Pre-training of deep bidirectional transformers for language understanding.
\newblock {\em arXiv preprint arXiv:1810.04805}, 2018.

\bibitem{raffel2020exploring}
Colin Raffel, Noam Shazeer, Adam Roberts, Katherine Lee, Sharan Narang, Michael Matena, Yanqi Zhou, Wei Li, and Peter~J Liu.
\newblock Exploring the limits of transfer learning with a unified text-to-text transformer.
\newblock {\em Journal of machine learning research}, 21(140):1--67, 2020.

\bibitem{touvron2023llama}
Hugo Touvron, Thibaut Lavril, Gautier Izacard, Xavier Martinet, Marie-Anne Lachaux, Timoth{\'e}e Lacroix, Baptiste Rozi{\`e}re, Naman Goyal, Eric Hambro, Faisal Azhar, et~al.
\newblock Llama: Open and efficient foundation language models.
\newblock {\em arXiv preprint arXiv:2302.13971}, 2023.

\bibitem{zellers2019defending}
Rowan Zellers, Ari Holtzman, Hannah Rashkin, Yonatan Bisk, Ali Farhadi, Franziska Roesner, and Yejin Choi.
\newblock Defending against neural fake news.
\newblock {\em Advances in neural information processing systems}, 32, 2019.

\bibitem{huang2022faking}
Kung-Hsiang Huang, Kathleen McKeown, Preslav Nakov, Yejin Choi, and Heng Ji.
\newblock Faking fake news for real fake news detection: Propaganda-loaded training data generation.
\newblock {\em arXiv preprint arXiv:2203.05386}, 2022.

\bibitem{chen2023can}
Canyu Chen and Kai Shu.
\newblock Can llm-generated misinformation be detected?
\newblock {\em arXiv preprint arXiv:2309.13788}, 2023.

\bibitem{su2023fake}
Jinyan Su, Terry~Yue Zhuo, Jonibek Mansurov, Di~Wang, and Preslav Nakov.
\newblock Fake news detectors are biased against texts generated by large language models.
\newblock {\em arXiv preprint arXiv:2309.08674}, 2023.

\bibitem{stewart2023efficacy}
Jake Stewart, Nikita Lyubashenko, and George Stefanek.
\newblock The efficacy of detecting ai-generated fake news using transfer learning.
\newblock {\em Issues in Information Systems}, 24(2), 2023.

\bibitem{jiang2023disinformation}
Bohan Jiang, Zhen Tan, Ayushi Nirmal, and Huan Liu.
\newblock Disinformation detection: An evolving challenge in the age of llms.
\newblock {\em arXiv preprint arXiv:2309.15847}, 2023.

\bibitem{papineni2002bleu}
Kishore Papineni, Salim Roukos, Todd Ward, and Wei-Jing Zhu.
\newblock Bleu: a method for automatic evaluation of machine translation.
\newblock In {\em Proceedings of the 40th annual meeting of the Association for Computational Linguistics}, pages 311--318, 2002.

\bibitem{lin2004rouge}
Chin-Yew Lin.
\newblock Rouge: A package for automatic evaluation of summaries.
\newblock In {\em Text summarization branches out}, pages 74--81, 2004.

\bibitem{zhang2019bertscore}
Tianyi Zhang, Varsha Kishore, Felix Wu, Kilian~Q Weinberger, and Yoav Artzi.
\newblock Bertscore: Evaluating text generation with bert.
\newblock {\em arXiv preprint arXiv:1904.09675}, 2019.

\bibitem{snover2006study}
Matthew Snover, Bonnie Dorr, Richard Schwartz, Linnea Micciulla, and John Makhoul.
\newblock A study of translation edit rate with targeted human annotation.
\newblock In {\em Proceedings of the 7th Conference of the Association for Machine Translation in the Americas: Technical Papers}, pages 223--231, 2006.

\bibitem{lee2023survey}
Seungjun Lee, Jungseob Lee, Hyeonseok Moon, Chanjun Park, Jaehyung Seo, Sugyeong Eo, Seonmin Koo, and Heuiseok Lim.
\newblock A survey on evaluation metrics for machine translation.
\newblock {\em Mathematics}, 11(4):1006, 2023.

\bibitem{lo2017meant}
Chi-kiu Lo.
\newblock Meant 2.0: Accurate semantic mt evaluation for any output language.
\newblock In {\em Proceedings of the second conference on machine translation}, pages 589--597, 2017.

\bibitem{mckeown1983paraphrasing}
Kathleen McKeown.
\newblock Paraphrasing questions using given and new information.
\newblock {\em American Journal of Computational Linguistics}, 9(1):1--10, 1983.

\bibitem{hassan2007unt}
Samer Hassan, Andras Csomai, Carmen Banea, Ravi Sinha, and Rada Mihalcea.
\newblock Unt: Subfinder: Combining knowledge sources for automatic lexical substitution.
\newblock In {\em Proceedings of the fourth international workshop on semantic evaluations (SemEval-2007)}, pages 410--413, 2007.

\bibitem{li2017paraphrase}
Zichao Li, Xin Jiang, Lifeng Shang, and Hang Li.
\newblock Paraphrase generation with deep reinforcement learning.
\newblock {\em arXiv preprint arXiv:1711.00279}, 2017.

\bibitem{gupta2018deep}
Ankush Gupta, Arvind Agarwal, Prawaan Singh, and Piyush Rai.
\newblock A deep generative framework for paraphrase generation.
\newblock In {\em Proceedings of the aaai conference on artificial intelligence}, volume~32, 2018.

\bibitem{hegde2020unsupervised}
Chaitra Hegde and Shrikumar Patil.
\newblock Unsupervised paraphrase generation using pre-trained language models.
\newblock {\em arXiv preprint arXiv:2006.05477}, 2020.

\bibitem{prakash2016neural}
Aaditya Prakash, Sadid~A Hasan, Kathy Lee, Vivek Datla, Ashequl Qadir, Joey Liu, and Oladimeji Farri.
\newblock Neural paraphrase generation with stacked residual lstm networks.
\newblock {\em arXiv preprint arXiv:1610.03098}, 2016.

\bibitem{he2016deep}
Kaiming He, Xiangyu Zhang, Shaoqing Ren, and Jian Sun.
\newblock Deep residual learning for image recognition.
\newblock In {\em Proceedings of the IEEE conference on computer vision and pattern recognition}, pages 770--778, 2016.

\bibitem{fu2019paraphrase}
Yao Fu, Yansong Feng, and John~P Cunningham.
\newblock Paraphrase generation with latent bag of words.
\newblock {\em Advances in Neural Information Processing Systems}, 32, 2019.

\bibitem{goyal2020neural}
Tanya Goyal and Greg Durrett.
\newblock Neural syntactic preordering for controlled paraphrase generation.
\newblock {\em arXiv preprint arXiv:2005.02013}, 2020.

\bibitem{li2019decomposable}
Zichao Li, Xin Jiang, Lifeng Shang, and Qun Liu.
\newblock Decomposable neural paraphrase generation.
\newblock {\em arXiv preprint arXiv:1906.09741}, 2019.

\bibitem{yu2022generate}
Wenhao Yu, Dan Iter, Shuohang Wang, Yichong Xu, Mingxuan Ju, Soumya Sanyal, Chenguang Zhu, Michael Zeng, and Meng Jiang.
\newblock Generate rather than retrieve: Large language models are strong context generators.
\newblock {\em arXiv preprint arXiv:2209.10063}, 2022.

\bibitem{witteveen2019paraphrasing}
Sam Witteveen and Martin Andrews.
\newblock Paraphrasing with large language models.
\newblock {\em arXiv preprint arXiv:1911.09661}, 2019.

\bibitem{wahle2022large}
Jan~Philip Wahle, Terry Ruas, Frederic Kirstein, and Bela Gipp.
\newblock How large language models are transforming machine-paraphrased plagiarism.
\newblock {\em arXiv preprint arXiv:2210.03568}, 2022.

\bibitem{yadav2024pag}
Vikas Yadav, Zheng Tang, and Vijay Srinivasan.
\newblock Pag-llm: Paraphrase and aggregate with large language models for minimizing intent classification errors.
\newblock In {\em Proceedings of the 47th International ACM SIGIR Conference on Research and Development in Information Retrieval}, pages 2569--2573, 2024.

\bibitem{jiao2019tinybert}
Xiaoqi Jiao, Yichun Yin, Lifeng Shang, Xin Jiang, Xiao Chen, Linlin Li, Fang Wang, and Qun Liu.
\newblock Tinybert: Distilling bert for natural language understanding.
\newblock {\em arXiv preprint arXiv:1909.10351}, 2019.

\bibitem{vogel2022investigating}
Liane Vogel and Lucie Flek.
\newblock Investigating paraphrasing-based data augmentation for task-oriented dialogue systems.
\newblock In {\em International Conference on Text, Speech, and Dialogue}, pages 476--488. Springer, 2022.

\bibitem{kumar2019submodular}
Ashutosh Kumar, Satwik Bhattamishra, Manik Bhandari, and Partha Talukdar.
\newblock Submodular optimization-based diverse paraphrasing and its effectiveness in data augmentation.
\newblock In {\em Proceedings of the 2019 Conference of the North American Chapter of the Association for Computational Linguistics: Human Language Technologies, Volume 1 (Long and Short Papers)}, pages 3609--3619, 2019.

\bibitem{mi2022improving}
Chenggang Mi, Lei Xie, and Yanning Zhang.
\newblock Improving data augmentation for low resource speech-to-text translation with diverse paraphrasing.
\newblock {\em Neural Networks}, 148:194--205, 2022.

\bibitem{patil2022t5w}
Annapurna~P Patil, Shreekant Jere, Reshma Ram, and Shruthi Srinarasi.
\newblock T5w: A paraphrasing approach to oversampling for imbalanced text classification.
\newblock In {\em 2022 IEEE International Conference on Electronics, Computing and Communication Technologies (CONECCT)}, pages 1--6. IEEE, 2022.

\bibitem{wieting2017paranmt}
John Wieting and Kevin Gimpel.
\newblock Paranmt-50m: Pushing the limits of paraphrastic sentence embeddings with millions of machine translations.
\newblock {\em arXiv preprint arXiv:1711.05732}, 2017.

\bibitem{dolan2005automatically}
Bill Dolan and Chris Brockett.
\newblock Automatically constructing a corpus of sentential paraphrases.
\newblock In {\em Third international workshop on paraphrasing (IWP2005)}, 2005.

\bibitem{ganitkevitch2013ppdb}
Juri Ganitkevitch, Benjamin Van~Durme, and Chris Callison-Burch.
\newblock Ppdb: The paraphrase database.
\newblock In {\em Proceedings of the 2013 conference of the north american chapter of the association for computational linguistics: Human language technologies}, pages 758--764, 2013.

\bibitem{dong2021parasci}
Qingxiu Dong, Xiaojun Wan, and Yue Cao.
\newblock Parasci: A large scientific paraphrase dataset for longer paraphrase generation.
\newblock {\em arXiv preprint arXiv:2101.08382}, 2021.

\bibitem{xu2013gathering}
Wei Xu, Alan Ritter, and Ralph Grishman.
\newblock Gathering and generating paraphrases from twitter with application to normalization.
\newblock In {\em Proceedings of the sixth workshop on building and using comparable corpora}, pages 121--128, 2013.

\bibitem{miao2019cgmh}
Ning Miao, Hao Zhou, Lili Mou, Rui Yan, and Lei Li.
\newblock Cgmh: Constrained sentence generation by metropolis-hastings sampling.
\newblock In {\em Proceedings of the AAAI Conference on Artificial Intelligence}, volume~33, pages 6834--6842, 2019.

\bibitem{jin2021euca}
Weina Jin, Jianyu Fan, Diane Gromala, Philippe Pasquier, and Ghassan Hamarneh.
\newblock Euca: The end-user-centered explainable ai framework.
\newblock {\em arXiv preprint arXiv:2102.02437}, 2021.

\bibitem{ribeiro2016should}
Marco~Tulio Ribeiro, Sameer Singh, and Carlos Guestrin.
\newblock " why should i trust you?" explaining the predictions of any classifier.
\newblock In {\em Proceedings of the 22nd ACM SIGKDD international conference on knowledge discovery and data mining}, pages 1135--1144, 2016.

\bibitem{szczepanskinew}
M~Szczepanski, M~Pawlicki, R~Kozik, and M~Choras.
\newblock New explainability method for bert-based model in fake news detection. sci. rep. 11 (1), 23705 (2021).

\bibitem{lundberg2017unified}
Scott Lundberg.
\newblock A unified approach to interpreting model predictions.
\newblock {\em arXiv preprint arXiv:1705.07874}, 2017.

\bibitem{ayoub2021combat}
Jackie Ayoub, X~Jessie Yang, and Feng Zhou.
\newblock Combat covid-19 infodemic using explainable natural language processing models.
\newblock {\em Information Processing \& Management}, 58(4):102569, 2021.

\bibitem{reis2019explainable}
Julio~CS Reis, Andr{\'e} Correia, Fabr{\'\i}cio Murai, Adriano Veloso, and Fabr{\'\i}cio Benevenuto.
\newblock Explainable machine learning for fake news detection.
\newblock In {\em Proceedings of the 10th ACM conference on web science}, pages 17--26, 2019.

\bibitem{sundararajan2017axiomatic}
Mukund Sundararajan, Ankur Taly, and Qiqi Yan.
\newblock Axiomatic attribution for deep networks.
\newblock In {\em International conference on machine learning}, pages 3319--3328. PMLR, 2017.

\bibitem{shrikumar2017learning}
Avanti Shrikumar, Peyton Greenside, and Anshul Kundaje.
\newblock Learning important features through propagating activation differences.
\newblock In {\em International conference on machine learning}, pages 3145--3153. PMlR, 2017.

\bibitem{alvarez2017causal}
David Alvarez-Melis and Tommi~S Jaakkola.
\newblock A causal framework for explaining the predictions of black-box sequence-to-sequence models.
\newblock {\em arXiv preprint arXiv:1707.01943}, 2017.

\bibitem{neely2022song}
Michael Neely, Stefan~F Schouten, Maurits Bleeker, and Ana Lucic.
\newblock A song of (dis) agreement: Evaluating the evaluation of explainable artificial intelligence in natural language processing.
\newblock In {\em HHAI2022: Augmenting Human Intellect}, pages 60--78. IOS Press, 2022.

\bibitem{ghaeini2018interpreting}
Reza Ghaeini, Xiaoli~Z Fern, and Prasad Tadepalli.
\newblock Interpreting recurrent and attention-based neural models: a case study on natural language inference.
\newblock {\em arXiv preprint arXiv:1808.03894}, 2018.

\bibitem{gholizadeh2021model}
Shafie Gholizadeh and Nengfeng Zhou.
\newblock Model explainability in deep learning based natural language processing.
\newblock {\em arXiv preprint arXiv:2106.07410}, 2021.

\bibitem{patwa2020fighting}
Parth Patwa, Shivam Sharma, Srinivas PYKL, Vineeth Guptha, Gitanjali Kumari, Md~Shad Akhtar, Asif Ekbal, Amitava Das, and Tanmoy Chakraborty.
\newblock Fighting an infodemic: Covid-19 fake news dataset, 2020.

\bibitem{wang2017liar}
William~Yang Wang.
\newblock ``liar, liar pants on fire'': A new benchmark dataset for fake news detection.
\newblock {\em arXiv preprint arXiv:1705.00648}, 2017.

\bibitem{loper2002nltk}
Edward Loper and Steven Bird.
\newblock Nltk: The natural language toolkit.
\newblock {\em arXiv preprint cs/0205028}, 2002.

\bibitem{gasparetto2022survey}
Andrea Gasparetto, Matteo Marcuzzo, Alessandro Zangari, and Andrea Albarelli.
\newblock A survey on text classification algorithms: From text to predictions.
\newblock {\em Information}, 13(2):83, 2022.

\bibitem{zhang2020pegasus}
Jingqing Zhang, Yao Zhao, Mohammad Saleh, and Peter Liu.
\newblock Pegasus: Pre-training with extracted gap-sentences for abstractive summarization.
\newblock In {\em International conference on machine learning}, pages 11328--11339. PMLR, 2020.

\bibitem{pedregosa2011scikit}
Fabian Pedregosa, Ga{\"e}l Varoquaux, Alexandre Gramfort, Vincent Michel, Bertrand Thirion, Olivier Grisel, Mathieu Blondel, Peter Prettenhofer, Ron Weiss, Vincent Dubourg, et~al.
\newblock Scikit-learn: Machine learning in python.
\newblock {\em Journal of machine learning research}, 12(Oct):2825--2830, 2011.

\bibitem{Taghizadeh2023}
Mohammad Taghizadeh.
\newblock flan-t5-base-imdb-text-classification.
\newblock \url{https://github.com/M-Taghizadeh/flan-t5-base-imdb-text-classification}, 2023.

\bibitem{yuan10distilbert}
Lik~Xun Yuan.
\newblock distilbert-base-multilingual-cased-sentiments-student (revision 2e33845), 2023.
\newblock {\em URL: https://huggingface. co/lxyuan/distilbert-base-multilingual-cased-sentiments-student. doi}, 10.

\end{thebibliography}

%TC:ignore
%\appendix

%\section{Example Appendix}
%\label{sec:appendix}

%This is an appendix.

%TC:endignore
\end{document}